\documentclass[a4paper]{llncs}

\usepackage{amssymb}
\usepackage{amsmath}

\usepackage{tikzexternal}
\usepackage{graphicx}
\usepackage{url}
\usepackage[latin1]{inputenc} 

\urldef{\mailsa}\path|{gabriel.kronberger, michael.kommenda, michael.affenzeller}@fh-hagenberg.at|
\urldef{\mailsb}\path|stefan.fink@jku.at|

\newcommand{\keywords}[1]{\par\addvspace\baselineskip
\noindent\keywordname\enspace\ignorespaces#1}

\begin{document}

\mainmatter  % start of an individual contribution

\title{Macro-Economic Time Series Modeling and Interaction Networks\footnote{The final publication is available at \url{http://link.springer.com/chapter/10.1007/978-3-642-20520-0_11}}}

%\titlerunning{Lecture Notes in Computer Science: Authors' Instructions}

\author{Gabriel Kronberger\inst{1}%
%\thanks{Heureka!, Austrian Research Promotion Agency, }%
\and Stefan Fink\inst{2}%
\and Michael Kommenda\inst{1} \and\\
Michael Affenzeller\inst{1}}
\institute{University of Applied Sciences Upper Austria, Research Center Hagenberg,\\
Softwarepark 11, 4232 Hagenberg, Austria\\
\mailsa\\
\and
Johannes Kepler University, Department of Economics\\
Altenbergerstraße 69, 4040 Linz, Austria\\
\mailsb\\
}
\authorrunning{Gabriel Kronberger, et al.}

\toctitle{Lecture Notes in Computer Science}
\tocauthor{Gabriel Kronberger, (Upper Austria University of Applied Sciences),
Stefan Fink, (Johannes Kepler University, Linz),
Michael Kommenda, (Upper Austria University of Applied Sciences),
Michael Affenzeller, (Upper Austria University of Applied Sciences)}

\maketitle

\begin{abstract}
Macro-economic models describe the dynamics of economic quantities. The estimations and forecasts produced by such models play a substantial role for financial and political decisions. In this contribution we describe an approach based on genetic programming and symbolic regression to identify variable interactions in large datasets. In the proposed approach multiple symbolic regression runs are executed for each variable of the dataset to find potentially interesting models. The result is a variable interaction network that describes which variables are most relevant for the approximation of each variable of the dataset.
This approach is applied to a macro-economic dataset with monthly observations of important economic indicators in order to identify potentially interesting dependencies of these indicators. The resulting interaction network of macro-economic indicators is briefly discussed and two of the identified models are presented in detail. The two models approximate the help wanted index and the CPI inflation in the US.
\end{abstract}

\section{Motivation}
Macro-economic models describe the dynamics of economic quantities of countries or regions, as well as their interaction on international markets. Macro-economic variables that play a role in such models are for instance the unemployment rate, gross domestic product, current account figures and monetary aggregates. Macro-economic models can be used to estimate the current economic conditions and to forecast economic developments and trends. Therefore macro-economic models play a substantial role in financial and political decisions. 

It has been shown by Koza that genetic programming can be used for econometric modeling \cite{Koza1990}, \cite{Koza1992}. He used a symbolic regression approach to rediscover the well-known exchange equation relating money supply, price level, gross national product and velocity of money in an economy, from observations of these variables. 

Genetic programming is an evolutionary method imitating aspects of biological evolution to find a computer program that solves a given problem through gradual evolutionary changes starting from an initial population of random programs \cite{Koza1992}. Symbolic regression is the application of genetic programming to find regression models represented as symbolic mathematical expressions. Symbolic regression is especially effective if little or no information is available about the studied system or process, because genetic programming is capable to evolve the necessary structure of the model in combination with the parameters of the model. 

In this contribution we take up the idea of using symbolic regression to generate models describing macro-economic interactions based on observations of economic quantities. However, contrary to the constrained situation studied in \cite{Koza1990}, we use a more extensive dataset with observations of many different economic quantities, and aim to identify all potentially interesting economic interactions that can be derived from the observations in the dataset. 
In particular, we describe an approach using GP and symbolic regression to generate a high level overview of variable interactions that can be visualized as a graph.

 Our approach is based on a large collection of diverse symbolic regression models for each variable of the dataset. In the symbolic regression runs the most relevant input variables to approximate each target variable are determined. This information is aggregated over all runs and condensed to a graph of variable interactions providing a coarse grained high level overview of variable interactions.

We have applied this approach on a dataset with monthly observations of economic quantities to identify (non-linear) interactions of macro-economic variables. 

\section{Modeling Approach}
The main objective discussed in this contribution is the identification of all potentially interesting models describing variable relations in a dataset. This is a broader aim than usually followed in a regression approach. Typically, modeling concentrates on a specific variable of interest (target variable) for which an approximation model is sought. Our aim resembles the aim of data mining, where the variable of interest is often not known a-priori and instead all quantities are analyzed in order to find potentially interesting patterns \cite{Hand2001}. 

\subsection{Comprehensive Symbolic Regression}
A straight forward way to find all potentially interesting models in a data set is to execute independent symbolic regression runs for all variables of the dataset building a large collection of symbolic regression models. This approach of comprehensive symbolic regression over the whole dataset is also followed in this contribution. 

Especially in real world scenarios there are often dependencies between the observed variables. In symbolic regression the model structure is evolved freely, so any combination of input variables can be used to model a given target variable. Even if all input variables are independent, a given function can be expressed in multiple different ways which are all semantically identical. This fact makes the interpretation of symbolic regression models difficult as each run produces a structurally different result. If the input variables are not independent, for instance a variable $x$ can be described by a combination of two other variables $y and z$, this problem is emphasized, because it is possible to express semantically equivalent functions using differing sets of input variables. 
A  benefit of the comprehensive symbolic regression approach is that dependencies of all variables in the dataset are made explicit in form of separate regression models. When regression models for dependencies of input variables are known, it is possible to detect alternative representations.

Collecting models from multiple symbolic regression runs is simple, but it is difficult to detect the actually interesting models \cite{Hand2001}. We do not discuss interestingness measures in this contribution. Instead, we propose a hierarchical approach for the analysis of results of multiple symbolic regression runs. On a high level, only aggregated information about relevant input variables for each target variable is visualized in form of a variable interaction network. If a specific variable interaction seems interesting, the models which represent the interaction can be analyzed in detail. 

Information about relevant variable interactions is implicitly contained in the symbolic regression models and distributed over all models in the collection. In the next section we discuss variable relevance metrics for symbolic regression which can be used to determine the relevant input variables for the approximation of a target variable.

\subsection{Variable Relevance Metrics for Symbolic Regression}
Information about the set of input variables necessary to describe a given dependent variable is often valuable for domain experts. For linear regression modeling, powerful methods have been described to detect the relevant input variables through variable selection or shrinkage methods \cite{Hastie2009}. However, if non-linear models are necessary then variable selection is more difficult. It has been shown that genetic programming implicitly selects relevant variables \cite{Langdon2004} for symbolic regression. Thus, symbolic regression can be used to determine relevant input variables even in situations where non-linear models are necessary. 

A number of different variable relevance metrics for symbolic regression have been proposed in the literature \cite{Vladislavleva2010b}.
In this contribution a simple frequency-based variable relevance metric is proposed, that is based on the number of variable references in all solution candidates visited in a GP run.

\subsection{Frequency-based Variable Relevance Metric}
The function $\text{relevance}_\text{freq}(x_i)$ is an indicator for the relative relevance of variable $x_i$. It is calculated as the average relative frequency of variable references $\text{freq}_\%(x_i, \text{Pop}_g)$ in population $\text{Pop}_g$ at generation $g$ over all $G$ generations of one run,
\begin{align}
  \text{relevance}_{\text{freq}}(x_i) & = \frac{1}{G} \sum_{g=1}^G \text{freq}_\%(x_i, \text{Pop}_g).
\label{equ_var_impact}
\end{align}

The relative frequency $\text{freq}_\%(x_i, \text{Pop})$ of variable $x_i$ in a population is the number of references $\text{freq}(x_i,\text{Pop})$ of variable $x_i$  over the number of all variable references,
\begin{align}
  \text{freq}_\%(x_i, \text{Pop}) & = \frac{\sum_{ s \in \text{Pop} } \text{RefCount}(x_i, s))}{\sum_{k=1}^n \sum_{ s \in \text{Pop} } \text{RefCount}(x_k, s)},
\end{align}
where the function $\text{RefCount}(x_i, s$) simply counts all references to variable $x_i$ in model $s$.

The advantage of calculating the variable relevance for the whole run instead of using only the last generation is that the dynamic behavior of variable relevance over the whole run is taken into account. The relevance of variables typically differs over multiple independent GP runs, because of the non-deterministic nature of the GP process. 
Therefore, the variable relevancies of one single GP run cannot be trusted fully as a specific variable might have a large relevance in a single run simply by chance. Thus, it is desirable to analyze variable relevance results over multiple GP runs in order to get statistically significant results. 

\section{Experiments}
We applied the comprehensive symbolic regression approach, described in the previous sections, to identify macro-economic variable interactions. In the following sections the macro-economic dataset and the experiment setup are described.

\subsection{Data Collection and Preparation}
\label{sec:dataset}
The dataset contains monthly observations of 33 economic variables and indexes from the United States of America, Germany and the Euro zone in the time span from 01/1980 -- 07/2007 (331 observations). The time series were downloaded from various sources and aggregated into one large  dataset without missing values.

Some of the time series in the dataset have a general rising trend and are thus also pairwise strongly correlated. The rising trend of these variables is not particularly interesting, so the derivatives (monthly changes) of the variables are studied instead of the absolute values. The derivative values ($d(x)$ in Figure \ref{figure:macro-economic-boxplot}) are calculated using the five point formula for the numerical approximation of the derivative \cite{NumericalRecipes2002} without prior smoothing. 

\subsection{Experiment Configuration}
The goal of the modeling step is to identify the network of relevant variable interactions in the macro-economic dataset. Thus, several symbolic regression runs were executed to produce approximation models for each variable as a function of the remaining 32 variables in the dataset. In this step symbolic regression models are generated for each of the 33 variables in separate GP runs. For each target variable 30 independent runs are executed to generate a set of different models for each variable.

The same parameter settings were used for all runs. Only the target variable and the list of allowed input variables were adapted. The GP parameter settings for our experiments are specified in Table \ref{table:gp-parameters-macro-economic}. We used rather standard GP configuration with tree-based solution encoding, tournament selection, sub-tree swapping crossover, and two mutation operators. The fitness function is the squared correlation coefficient of the model output and the actual values of target variables. Only the final model is linearly scaled to match the location and scale of the target variable \cite{Keijzer2004}. The function set includes arithmetic operators (division is not protected) and additionally symbols for the arithmetic mean, the logarithm function, the exponential function and the sine function. The terminal set includes random constants and all 33 variables of the dataset except for the target variable. The variable can be either non-lagged or lagged up to 12 time steps. All variables contained in the dataset are listed in Figures \ref{figure:macro-economic-boxplot} and\ref{figure:macro-economic-network}.

 Two recent adaptations of the algorithm are included to reduce bloat and overfitting. Dynamic depth limits \cite{Silva:2009:GPEM} with an initial depth limit of seven are used to reduce the amount of bloat. An internal validation set is used to reduce the chance of overfitting. Each solution candidate is evaluated on the training and on the validation set. Selection is based solely on the fitness on the training set; the fitness on the validation set is used as an indicator for overfitting. 
Models which have a high training fitness but low validation fitness are likely to be overfit. Thus, the Spearman's rank correlation $\rho(\text{Fitness}_\text{train},\text{Fitness}_\text{val})$ of training- and validation fitness of all solution candidates in the population is calculated after each generation. If the correlation of training- and validation fitness in the population drops below a certain threshold the algorithm is stopped.

The dataset has been split into two partitions; observations 1--300 are used for training, observations 300--331 are used as a test set. Only observations 13--200 are used for fitness evaluation, the remaining observations of the training set are used as internal validation set for overfitting detection and for the selection of the final (best on validation) model. 

\begin{table}
\centering
\small
\begin{tabular}{l|l}
Parameter & Value\\
\hline
Population size & 2000\\
Max. generations & 150\\
Parent selection & Tournament (group size = 7) \\
Replacement & 1-Elitism \\
Initialization & PTC2 \cite{luke:2000:2ftcaGP}\\ 
Crossover & Sub-tree-swapping\\
Mutation & 7\% One-point, 7\% sub-tree replacement\\
Tree constraints & Dynamic depth limit (initial limit = 7)\\
Model selection & Best on validation\\
Stopping criterion & $\rho(\text{Fitness}_\text{train},\text{Fitness}_\text{val})  < 0.2$\\
Fitness function & $R^2$ (maximization)\\
Function set & +, -, *, /, avg, log, exp, sin\\
Terminal set & constants, variables, lagged variables (t-12) \ldots (t-1)
\end{tabular}
\caption{Genetic programming parameters.}
\label{table:gp-parameters-macro-economic}
\end{table}

\section{Results}
\begin{figure}
\centering
\includegraphics[width=8cm]{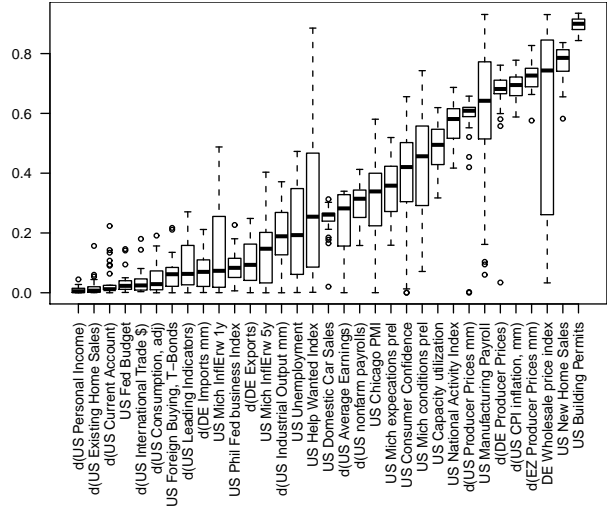}
\caption{Box-plot of the squared Pearson's correlation coefficient ($R^2$) of the model output and actual values on the test set for each variable.}
\label{figure:macro-economic-boxplot}
\end{figure}

For each variable of the dataset 30 independent GP runs have been executed using the open source software HeuristicLab. The result is a collection of 990 models, 30 symbolic regression models for each of the 33 variables generated in 990 GP runs. The collection of all models represents all identified (non-linear) interactions between all variables. 
Figure \ref{figure:macro-economic-boxplot} shows the box-plot of the squared Pearson's correlation coefficient ($R^2$) of the model output and the original values of the target variable on the test set for the 30 models for each variable. 

\subsection{Variable Interaction Network}
\begin{figure}
\centering
\includegraphics[width=10cm]{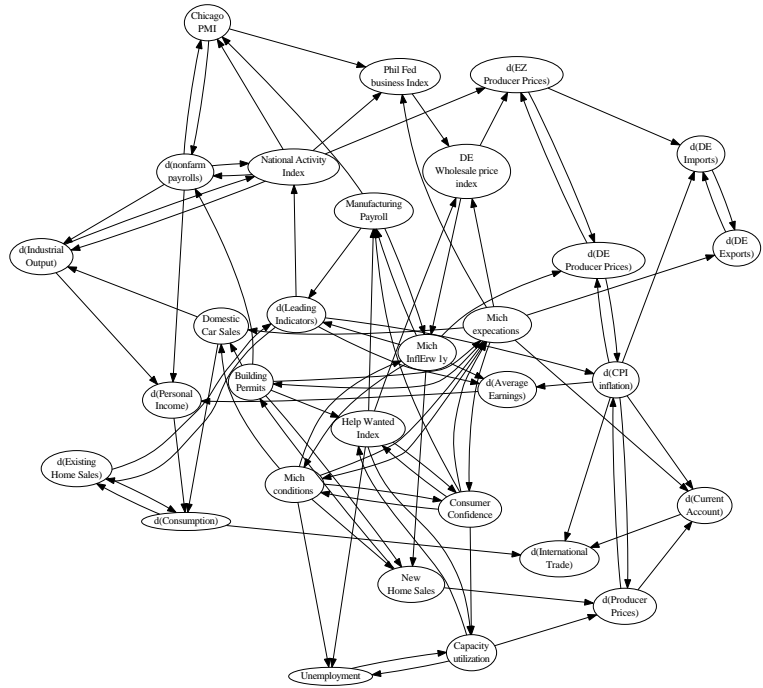}
\caption{Variable interaction network of macro-economic variables identified through comprehensive symbolic regression and frequency-based variable relevance metrics. This figure has been plotted using GraphViz.}
\label{figure:macro-economic-network}
\end{figure}

In Figure \ref{figure:macro-economic-network} the three most relevant input variables for each target variable are shown where an arrow ($a \rightarrow b$) means that variable $a$ is a relevant variable for modeling variable $b$. In the interaction network variable $a$ is connected to $b$ ($a \rightarrow b$) if $a$ is among the top three most relevant input variables averaged over all models for variable $b$, where the variable relevance is calculated using the metric shown in Equation \ref{equ_var_impact}. The top three most important input variables are determined for each of the 33 target variables in turn and GraphViz is used to layout the resulting network shown in Figure \ref{figure:macro-economic-network}.

The network of relevant variables shows many strong double-linked variable relations. GP discovered strongly related variables, for instance \textit{exports} and \textit{imports} of Germany, \textit{consumption} and \textit{existing home sales}, \textit{building permits} and \textit{new home sales}, \textit{Chicago PMI} and \textit{non-farm payrolls} and a few more. GP also discovered a chain strongly related variables connecting the \textit{producer price indexes} of the euro zone, Germany and the US with the \textit{US CPI inflation}. 

A large strongly connected cluster that contains the variables \textit{unemployment}, \textit{capacity utilization}, \textit{help wanted index}, \textit{consumer confidence}, \textit{U.Mich. expectations}, \textit{U.Mich. conditions}, \textit{U.Mich. 1-year inflation}, \textit{building permits}, \textit{new home sales}, and \textit{manufacturing payrolls} has also been identified by our approach. 

Outside of the central cluster the variables \textit{national activity index}, \textit{CPI inflation}, \textit{non-farm payrolls} and \textit{leading indicators} also have a large number of outgoing connections indicating that these variables play an important role for the approximation of many other variables. 

\subsection{Detailed Models}
The variable interaction network only provides a course grained high level view on the identified macro-economic interactions. To obtain a better understanding of the identified macro-economic relations it is necessary to analyze single models in more detail. Because of space constraints we cannot give a full list of the best model identified for each variable in the data set. We selected two models for the \textit{Help wanted index} and \textit{CPI inflation} instead, which are discussed in more detail in the following sections.

The help wanted index is calculated from the number of job advertisements in major newspapers and is usually considered to be related to the unemployment rate \cite{Cohen1967}, \cite{Abraham1987}. 
The model for the \textit{help wanted index} shown in Equation \ref{eqn:help-wanted-index}  has a $R^2$ value of 0.82 on the test set. The model has been simplified manually and constant factors are not shown to improve comprehensibility. The model includes the \textit{manufacturing payrolls} and the \textit{capacity utilization} as relevant factors. Interestingly, the unemployment rate which was also available as input variable is not used, instead other indicators for economic conditions (\textit{Chicago PMI}, \textit{U. Mich cond.}) are included in the model. Interestingly the model also includes the \textit{building permits} and \textit{wholesale price index of Germany}. 
% R^2 0.82
\begin{align}
\begin{split}
\text{Help wanted index} & = {} \text{Building permits}  +  \text{Mfg payroll} +  \text{Mfg Payroll} (t-5)\\
 &  +  \text{Capacity utilization}  +  \text{Wholesale price index (GER)}\\
 &  +  \text{Chicago PMI} +\text{U. Mich cond.} (t-3)  
\end{split}
\label{eqn:help-wanted-index}
\end{align}

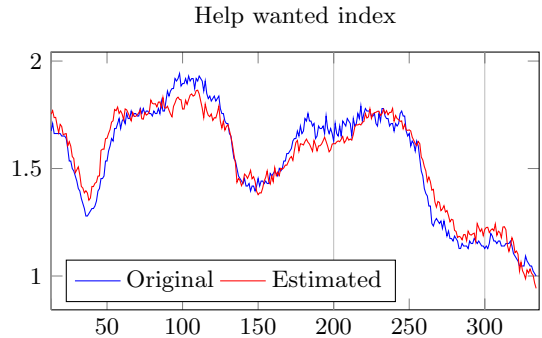
\begin{figure}
\centering
\begin{tikzpicture}
\pgfplotstableread[
  col sep=tab, 
  row sep=newline,
]
{help_wanted_index-output.txt}\fullresultstable
\begin{axis}[
   title = {Help wanted index},
   legend pos = {south west},
   legend columns=-1,
   legend entries = {Original, Estimated },
   width = 8cm,
   height = 5cm, 
   xmin=13,
   xmax=336,
   no markers,  
   extra x ticks={13, 200, 300},
   extra x tick style={grid=major},
   extra x tick labels={},
   ]
  \addplot table[x=Index, y=Original] \fullresultstable;
  \addplot table[x=Index, y=Estimated] \fullresultstable;
\end{axis}
\end{tikzpicture}
\caption{Line chart of the actual value of the \textit{US Help wanted index} and the estimated values produced by the model (Equation \ref{eqn:help-wanted-index}). Test set starts at index 300.}
\label{figure:help-wanted-index}
\end{figure}

Figure \ref{figure:help-wanted-index} shows a line chart for the actual values of the \textit{help wanted index} in the US and the estimated values of the model (Equation \ref{eqn:help-wanted-index}) over the whole time span covered by the dataset.

The \textit{consumer price index} measures the change in prices paid by customers for a certain market basket containing goods and services, and is measure for the inflation in an economy. The output of the model for the \textit{CPI inflation} in the US shown in Equation \ref{eqn:cpi-inflation} is very accurate with a squared correlation coefficient of 0.93 on the test set. This model has also been simplified manually and again constant factors are not shown to improve comprehensibility.  The model approximates the \textit{consumer price index} based on the \textit{unemployment}, \textit{car sales}, \textit{New home sales}, and the \textit{consumer confidence}. 
% 0.93
\begin{align}
\begin{split}
\text{CPI inflation}  & = {} \text{Unemployment}  +  \text{Domestic car sales} + \text{New home sales}\\
 & + \log(\text{New home sales} (t-4)  +  \text{New home sales} (t-2)\\
 & \qquad + \text{Consumer conf.}(t-1) +  \text{Unemployment} (t-5))
\end{split}
\label{eqn:cpi-inflation}
\end{align}

\begin{figure}
\centering
\begin{tikzpicture}
\pgfplotstableread[
  col sep=tab, 
  row sep=newline,
]
{cpi_inflation-output.txt}\fullresultstable
\begin{axis}[
   title = {CPI inflation},
   legend pos = {north west},
   legend columns=-1,
   legend entries = {Original, Estimated },
   width = 8cm,
   height = 5cm, 
   xmin=13,
   xmax=336,
   no markers,  
   extra x ticks={13, 200, 300},
   extra x tick style={grid=major},
   extra x tick labels={},
   ]
  \addplot table[x=Index, y=Original] \fullresultstable;
  \addplot table[x=Index, y=Estimated] \fullresultstable;
\end{axis}
\end{tikzpicture}
\caption{Line chart of the actual value of the \textit{US CPI inflation} and the estimated values produced by the model (Equation \ref{eqn:cpi-inflation}).}
\label{fig:cpi-inflation}
\end{figure}
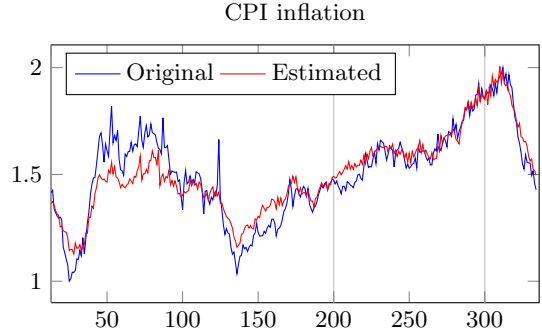

Figure \ref{fig:cpi-inflation} shows a line chart for the actual values of the \textit{CPI inflation} in the US and the estimated values of the model (Equation \ref{eqn:cpi-inflation}) over the whole time span covered by the dataset. Notably the drop of the CPI in the test set (starting at index 300) is estimated correctly by the model.

\section{Conclusion}
The application of the proposed approach on the macro-economic dataset resulted in a high level overview of macro-economic variable interactions.
In the experiments we used dynamic depth limits to counteract bloat and an internal validation set to detect overfitting using the correlation of training- and validation fitness. Two models for the \textit{US Help wanted index} and the \textit{US CPI inflation} have been presented and discussed in detail. Both models are rather accurate also on the test set and are relatively comprehensible.

We suggest using this approach for the exploration of variable interactions in a dataset when approaching a complex modeling task. The visualization of variable interaction networks can be used to give a quick overview of the most relevant interactions in a dataset and can help to identify new unknown interactions. The variable interaction network provides information that is not apparent from analysis of single models, and thus supplements the information gained from detailed analysis of single models.

\subsubsection*{Acknowledgments} 
This work mainly reflects research work done within the Josef Ressel-center for heuristic optimization ``Heureka!'' at the Upper Austria University of Applied Sciences, Campus Hagenberg. The center ``Heureka!'' is supported by the Austrian Research Promotion Agency (FFG) on behalf of the Austrian Federal Ministry of Economy, Family and Youth (BMWFJ).

\bibliographystyle{splncs03}
\bibliography{kronberger}

\end{document}